\documentclass[mlabstract]{jmlr}

\usepackage{wrapfig}

\usepackage{longtable}

\usepackage{booktabs}
\usepackage[load-configurations=version-1]{siunitx} 

\theorembodyfont{\upshape}
\theoremheaderfont{\scshape}
\theorempostheader{:}
\theoremsep{\newline}

\usepackage{amsmath}
\usepackage{amssymb}
\usepackage{amsfonts}
\jmlrvolume{}
\firstpageno{1}

\jmlryear{2024}
\jmlrworkshop{Symmetry and Geometry in Neural Representations}

\title[An Analysis on Cross-Dataset Transfer of Pretrained GNNs]{Towards Foundation Models on Graphs: An Analysis on Cross-Dataset Transfer of Pretrained GNNs}

\author{\Name{Fabrizio Frasca} \hfill \Name{Fabian Jogl} \\
\addr Technion – Israel Institute of Technology \hfill TU Wien \\
\Email{fabriziof@campus.technion.ac.il} \hfill \Email{fabian.jogl@tuwien.ac.at}
\AND
\Name{Moshe Eliasof} \hfill \Name{Matan Ostrovsky} \\
\addr University of Cambridge \hfill Technion – Israel Institute of Technology
\AND
\Name{Carola-Bibiane Schönlieb} \hfill \Name{Thomas Gärtner} \\
\addr University of Cambridge \hfill TU Wien 
\AND
\Name{Haggai Maron} \\
\addr Technion – Israel Institute of Technology, NVIDIA Research
}

\newcommand{\zinc}[0]{\texttt{zinc}}
\newcommand{\molpcba}[0]{\texttt{molpcba}}
\newcommand{\pept}[0]{\texttt{peptides}}
\newcommand{\zm}[0]{\texttt{zinc+molpcba}}

\newcommand{\molpep}[0]{\texttt{molpcba+peptides}}
\newcommand{\Q}[1]{(\textbf{Q#1})}

\begin{document}

\maketitle

\begin{abstract}
    To develop a preliminary understanding towards Graph Foundation Models, we study the extent to which pretrained Graph Neural Networks can be applied across datasets, an effort requiring to be agnostic to dataset-specific features and their encodings.
    We build upon a purely structural pretraining approach and propose an extension to capture feature information while still being feature-agnostic.
    We evaluate pretrained models on downstream tasks for varying amounts of training samples and choices of pretraining datasets.
    Our preliminary results indicate that embeddings from pretrained models improve generalization only with enough downstream data points and in a degree which depends on the quantity and properties of pretraining data. Feature information can lead to improvements, but currently requires some similarities between pretraining and downstream feature spaces.
\end{abstract}


\section{Introduction}\label{sec:intro}

Machine Learning is being revolutionized by Foundation Models (FMs), reference models that can be effectively adapted to diverse downstream tasks with a relatively small amount of data. Successful FMs abound in the textual and visual domains, but the development of FMs on graphs is still an open problem~\citep{liu_towards_2023,mao_graph_2024} with a dangling fundamental question: \emph{How to design an architecture that can be pretrained and, later, successfully applied across graph datasets?} This is challenging, as graphs representing different objects may exhibit different patterns in their structure and features; these last ones, across domains, may even represent semantically different spaces with distinct encodings.

If anything, such an architecture is required to be \emph{feature-agnostic}: it may process features, but in a way that is independent of dataset-specific encodings and that encourages learning of transferable patterns. While researchers have recently started to develop feature-agnostic approaches, they are still plagued by various limitations (see \appendixref{apd:related-work}).

Towards a more foundational understanding of the matter, we find a constructive reference in the pipeline proposed by~\citet{canturk2024graph}. The authors propose GPSE, a method to \emph{learn} graph positional and structural encodings (P/SEs)~\citep{dwivedi2021graph,rampasek_recipe_2023} that can be employed to augment any downstream Graph Neural Network (GNN). It consists of an expressive GNN pretrained on featureless graphs to predict P/SEs. Given a downstream graph task, the model can generate encodings that can be combined with explicit node features and fed into a GNN architecture. While pretrained only on a single molecular dataset, the authors show GPSE encodings transfer to different downstream datasets. GPSE represents an important step towards feature-agnostic approaches, but the pretraining phase disregards features, which may constitute an important source of information. Also, the authors do not explore settings of interest for FMs, viz., multi-dataset pretraining and downstream low-data regimes, and do not discuss label-free approaches to validate a pretrained model for a downstream target dataset.

In this work, we analyze the aforementioned aspects. We propose \emph{Feature-Structuralization}\footnote{We may abbreviate this simply as ``\emph{structuralization}'' or ``\emph{struct}''.} to encode feature information in structural form in a way that is compatible with structural pretraining strategies. Next, we consider three datasets that differ in their structural and feature patterns. We analyze the behavior and performance of structuralization and the original GPSE model on them, studying the impact of multi-dataset pretraining and of feature information in successfully transferring to downstream low-data regimes.

\section{Feature-Structuralization}
\label{sec:structuralization}

\begin{wrapfigure}{r}{0.6\textwidth}
    \vspace{-1cm}
    \centering
    \includegraphics[width=0.49\linewidth]{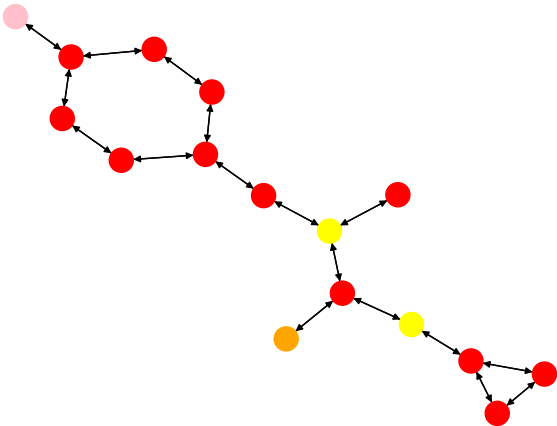}
    \includegraphics[width=0.49\linewidth]{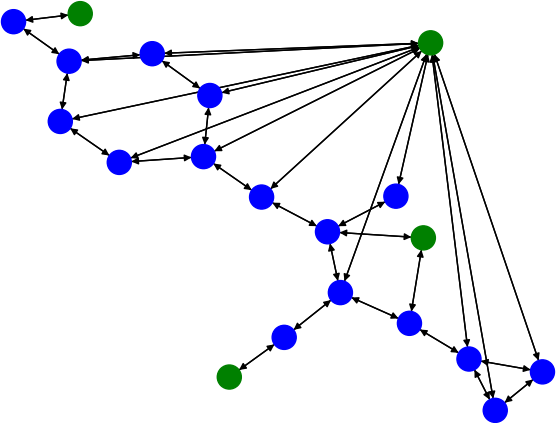}
    \caption{Left: graph from \zinc{}; Right: structuralized form. Colors represent node features.}
    \label{fig:struct_vis}
    \vspace{-0.5cm}
\end{wrapfigure}

Structuralization encodes feature information into additional structure augmenting the original input graph: by only encoding abstract, logical relations induced by features, the pretraining approach remains feature-agnostic while still accessing meaningful information on how features interact with the original graph structure.

Consider a graph $G$ with $d$ different categorical node features\footnote{Our approach can be extended to consider edge features and non-categorical ones. To represent edge features, one could consider an ``incidence network'' representation of the graph~\citep{albooyeh2019incidence}, where edges are effectively materialized into nodes. In that case, Feature-Structuralization would apply as described above. As for non-categorical (continuous) features, we can represent them as edge attributes in the structuralized form of the input graph. For this, we materialize a single feature-node with connections to all original nodes and encode the value of the continuous features as edge features thereon.}. For each channel $i = 1, \dots d$, structuralization materializes an additional \emph{feature-node} $u_{i,j}$ for any category $j$ in $i$, and connects $u_{i,j}$ to the nodes $v \in G$ attributed with category $j$ in channel $i$. In the transformed graph, all explicit features are discarded; nodes are only assigned a mark which allows to disambiguate between original and feature-nodes. An example of a single-channel structuralization is shown in~\figureref{fig:struct_vis} where, on the transformed graph, feature-nodes are in green, original nodes in blue.

\section{Experimental Setting}\label{sec:exp-set}

We seek to address the following questions: \Q{1} ``\textit{Are embeddings from pretrained models beneficial in downstream low-data regimes?}'' \Q{2} ``\textit{How does the composition of the pretraining corpus affect downstream generalization performance?}'' \Q{3} ``\textit{Is pretraining with features information beneficial when these are incorporated via structuralization?}''. We now describe our experimental setting and refer to \appendixref{apd:exp-set} for additional details.

\paragraph{Datasets.} We opt to use three datasets: ZINC-12k~\citep{dwivedi2020benchmarking}, ogbg-molpcba~\citep{hu2020open}, and peptides-func~\citep{dwivedi2022long}(resp.\ \zinc, \molpcba, \pept). These are from a similar domain, but with different degrees of similarity in structures and features: \pept's graphs are aminoacid-chains, structurally dissimilar from the graphs in \zinc\ and \molpcba\ (small molecules); nodes in \molpcba\ and \pept\ have the same $9$ SMILES-derived categories as features, nodes in \zinc\ only one category (the atom type).

\paragraph{Pretraining.} All pretrainings use the same base model from GPSE~\citep{canturk2024graph}, with either the original graph structures or their feature-structuralized variants. Following \citet{canturk2024graph}, pretraining involves jointly predicting predefined P/SEs from the output node representations. For structuralization, the model predicts P/SEs for the original nodes in both the original and transformed graphs. Pretraining is conducted on each dataset and on all pairwise combinations, totaling 12 possibilities. We downsampled \molpcba\ to match the training set sizes of \zinc\ and \pept.

\begin{wrapfigure}{r}{0.4\textwidth}
    \centering
    \includegraphics[width=0.99\linewidth]{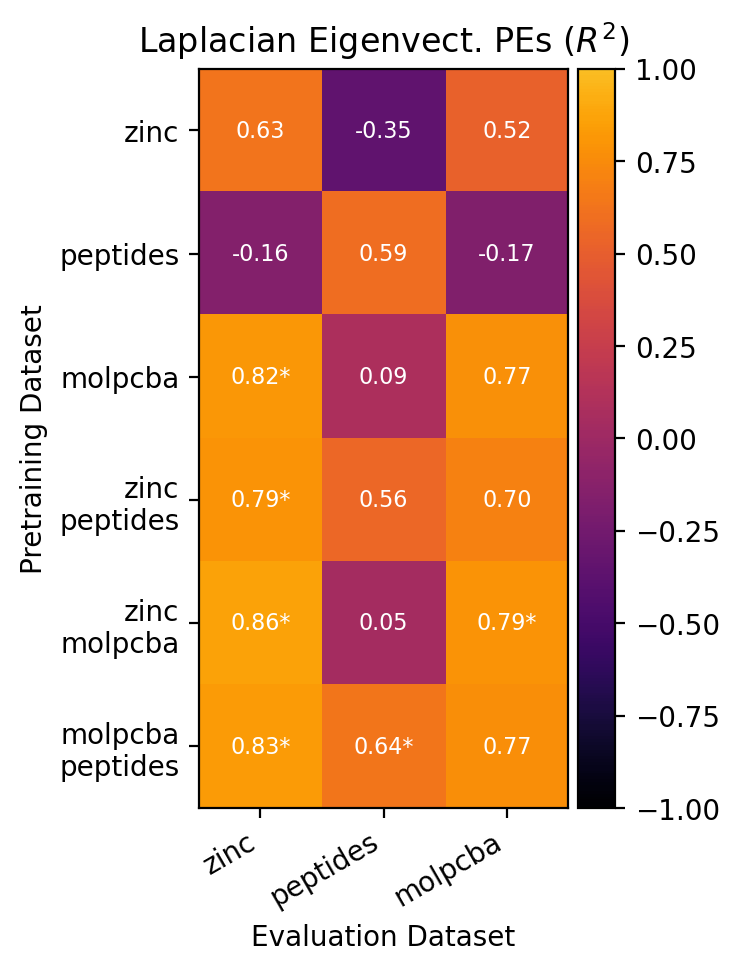}
    \caption{GPSE pretraining eval. on LapPEs (Test $R^2$).}
    \label{fig:pre_gpse_lappe}
    \vspace{-1.5cm}
\end{wrapfigure}

\paragraph{Evaluations.} We evaluate the pretrained models in two settings: (i) pretraining target prediction, assessing how well the models predict test targets from the three datasets; and (ii) downstream tasks, examining if and how the pretrained embeddings improve GNN generalization. These analyses are performed with varying downstream training sample sizes.

\section{Results and Discussion}\label{sec:disc}

\paragraph{Pretraining.} \figureref{fig:pre_gpse_lappe} visualizes how pretrained GPSEs generalize for one pretraining target (results for other targets can be found in Appendix~\ref{apd:res-pre}). This shows that pretraining on multiple datasets generally improves performance.\footnote{The pretraining amounts to a regression task. Performance is therefore measured in terms of the coefficient of determination $R^2$, consistently with previous works~\citep{canturk2024graph}.} Models not pretrained on \pept{} perform near to chance on \pept{}, unlike \zinc{} or \molpcba{}. While the gap between in- and off-domain pretraining narrows with more data, it never fully closes (see Appendix~\ref{apd:more-data}). Adding data from other sources can still be beneficial. Pretraining on ``structuralized graphs'' slightly worsens in-domain and hampers out-of-dataset generalization. 

\begin{figure}
    \centering
    \includegraphics[width=0.4\linewidth]{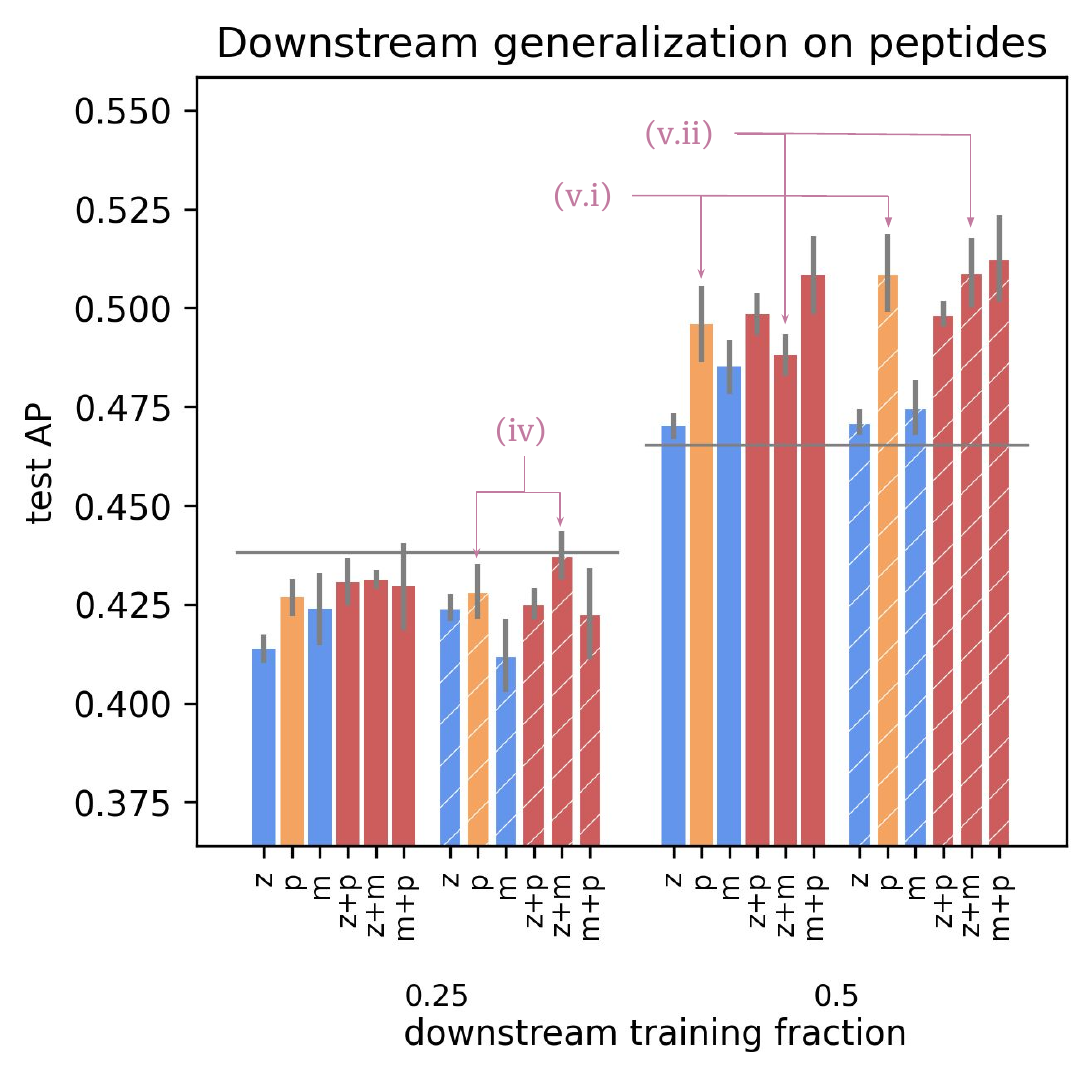}
    \includegraphics[width=0.5\linewidth]{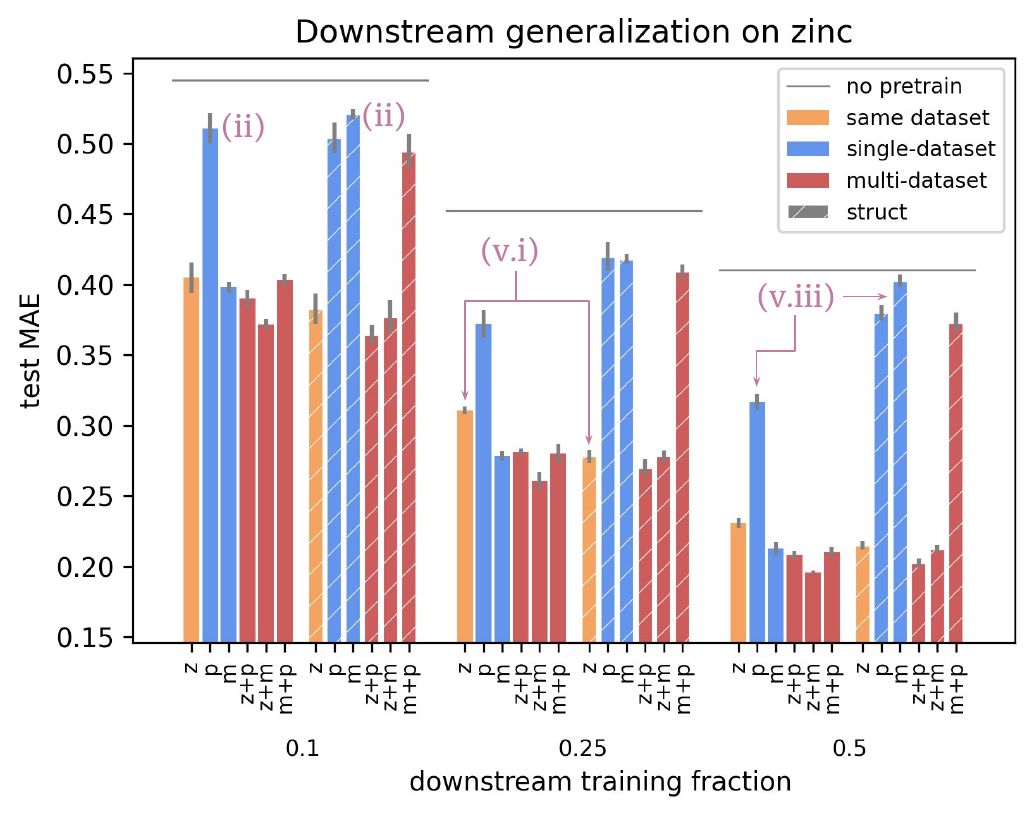}
    \caption{Downstream test performance for different training fraction and pretraining data (`z': \zinc, `p': \pept, `m': \molpcba). Left: Average Precision on \pept, higher is better. Right: Mean Absolute Error on \zinc, lower is better.}
    \label{fig:down}
    \vspace{-0.5cm}
\end{figure}

\paragraph{Downstream applications.}
We now present observations for downstream evaluations on \zinc\ and \pept, which we visually summarize in~\figureref{fig:down}. We also ran preliminary experiments on \molpcba, but noticed an inherent task hardness beyond the aforementioned datasets; our initial results are in~\appendixref{apd:res-down}, where we also report an extended version of~\figureref{fig:down}. Next, when referring to a `baseline', we intend the same downstream architecture ``deprived'' of embedding from pretrained models.
\\
\textbf{(i) Pre-trained embeddings can be --- but are not always --- beneficial} \Q{1}. Pretrained embeddings improve generalization over the baseline with sufficient data, but can be detrimental in data-scarce settings\footnote{The following observations focus on cases where pretraining is comparable or better than the baseline.}. Notably, in \zinc, pretrained models outperform the baseline even with just $10\%$ of the training data.  \\
\textbf{(ii) When pretraining becomes beneficial, all pretraining mixes seem to provide better generalization than the baseline} \Q{2}. However, the composition of the corpus may have a strong impact, which tend to reflect the similarity between source and target datasets. E.g., pretraining on \pept\ is less beneficial when transferring on \zinc\ for both GPSE and Struct, due to differences in both structure and features (see mark ii). \\
\textbf{(iii) Including datasets in pretraining other than the target does not generally deteriorate performance} \Q{2}. In general, the performance is at least as good as the one attained by the best of the two pretraining datasets. \\
\textbf{(iv) Pretraining mixes that do not include the target dataset can perform as well as pretraining on the target} \Q{2}.  In some cases, ``off-dataset'' pretraining may even generalize slightly better than pretraining on the same dataset (see mark iv). We believe that this may be mostly due to an increase in pretraining data (see~\appendixref{apd:more-data}). \\
\textbf{(v) Feature structuralization does not yield a consistent, significant improvement across settings} \Q{3}. Structuralization improves performance when in-dataset pretraining, outperforming vanilla GPSE on \zinc\ when pretrained on \zinc\ (mark v.i) and on \pept\ when pretrained on \pept\ (mark v.i). It can also help when the pretraining mix is off-distribution in structure but not features, e.g., on \pept\ with \zm\ pretraining (mark v.ii). However, structuralization is less effective when features are off-distribution, such as pretraining on \molpcba\ and transferring to \zinc\ (mark v.iii).

We refer to \appendixref{apd:more-data} for more considerations on the size of pretraining corpora.

\section{Conclusions}

We analyzed the extent to which pretrained GNNs can be transferred across datasets by measuring the impact of pretraining datasets on downstream generalization and the inclusion of feature information via structuralization. Our work can be extended in different ways by: studying non-molecular domains, e.g., social and collaboration networks; closely enquiring into the underperformance of structuralization and designing more compatible pretraining strategies and architectures; exploring other feature-agnostic pretraining approaches, e.g., to natively handle continuous features for domains such as geometric graphs.

\acks{

     The authors would like to thank Pascal Welke, David Penz, Sagar Malhotra, Marco Ciccone and Ethan Fetaya for helpful discussions.
     FF is funded by the Andrew and Erna Finci Viterbi Post-Doctoral Fellowship. FF performed this work while visiting the Machine Learning Research Unit at TU Wien led by Prof.\ Thomas Gärtner.
     FJ is funded by the Center for Artificial Intelligence and Machine Learning (CAIML) at TU Wien.
     ME is funded by the Blavatnik-Cambridge fellowship, the Cambridge Accelerate Programme for Scientific Discovery, and the Maths4DL EPSRC Programme.
    TG is supported by the Vienna Science and Technology Fund (WWTF) through project ICT22-059.
     HM is a Robert J.\ Shillman Fellow and is supported by the Israel Science Foundation through a personal grant (ISF 264/23) and an equipment grant (ISF 532/23). 
}

\bibliography{refs}
\clearpage
\newpage
\appendix

\section{Dealing with different features across datasets}\label{apd:related-work}

Pioneering approaches for feature-agnostic architectures have either resorted to (i) Large Language Models (LLMs) to embed features in a shared semantic space~\citep{liu_one_2023,huang_prodigy_2023}, (ii) adopted dimensionality reduction techniques~\citep{zhao2024all,yu2024text}, or (iii) assumed the architecture is parameterized by a `bank' of linear GNNs whose optimal predictions can be precalculated in closed form~\citep{zhao2024graphany}.

As for (i), end users have little control over the semantic relations between the LLM-encoded features, which can also significantly depend on the prompting pattern and technique. This is particularly relevant when working with `abstract' categorical features or continuous ones.
Techniques in class (ii) leverage covariance information in the features to derive a certain number of principal components. This can be interpreted as a form of `canonization', which, however, neglects structural information in the graph and does not necessarily guarantee, in itself, a form of alignment of the semantic spaces across datasets. Importantly, we note that covariance information is trivial for purely categorical feature spaces, e.g., following a one-hot encoding scheme.
Approach (iii) is particularly interesting as the learnable components of the approach effectively work in the prediction space, sidestepping the problem of aligning semantic spaces across datasets. However, it poses an architectural constraint, viz., the use of linear GNNs, that may be too limiting in applications such as the ones considered in this study.

Beyond our scope, in the specific context of Knowledge Graphs, \citet{galkin_towards_2023} propose to construct a graph between relation types based on concept of sharing `head' or `tail' entities. Message-passing on this graph allows the representation of any relation type and thus enables applications across Knowledge Graphs. Structuralization may share similarities with this approach in the high-level intuition; however, graphs generated by structuralization jointly represent original and feature-induced nodes and edges. The model being pretrained can access both and potentially (learn to) capture patterns in their interaction.

\section{More experimental details}\label{apd:exp-set}

\begin{figure}
    \centering
    \includegraphics[width=0.99\linewidth]{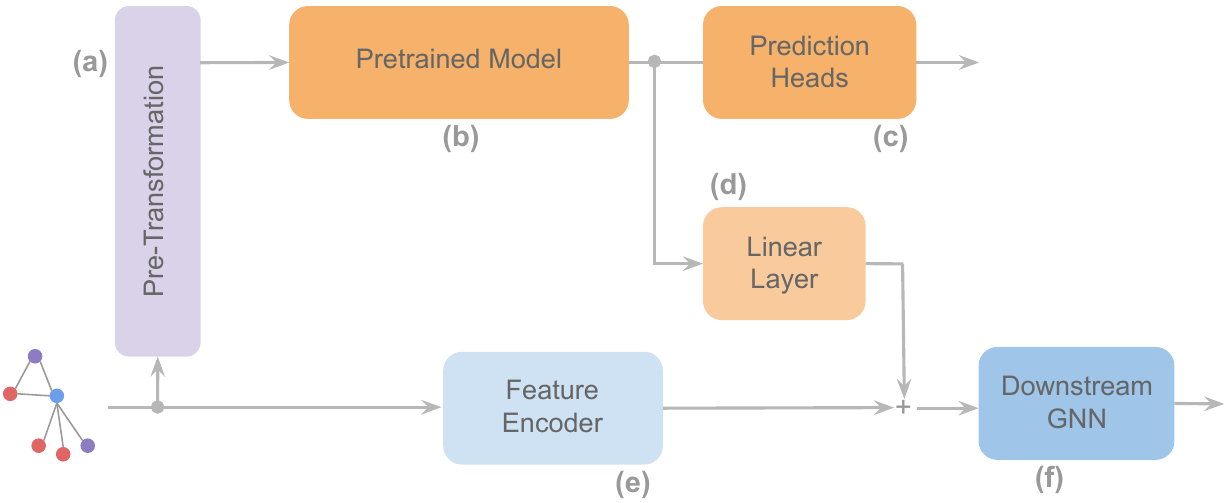}
    \caption{Architecure pipeline for downstream experiments.}
    \label{apd:fig:pipeline}
\end{figure}

\subsection{Pretraining and pretraining target prediction}

In all pretrainings, we employ the same backbone architecture proposed by~\citet{canturk2024graph}: a $20$-layer GNN with $512$-dimensional Residual Gated GCN message-passing layers~\cite{bresson2017residual} and prediction heads constituted by $2$-Layer Perceptrons with $32$ hidden-dimensions, totaling slightly more than $2$M parameters. The layers of the backbone architecture are regularized with dropout, whose rate is set to $0.2$. As in~\citep{canturk2024graph}, input graphs are augmented with virtual nodes, and $20$-dimensional random node features drawn from a standard Normal distribution.

For optimization, we follow the setting proposed in~\citep{canturk2024graph}: we employ the AdamW optimizer~\citep{loshchilov2019decoupled} for $120$ epochs with weight decay set to $0.00001$, a base learning rate of $0.005$ and a Cosine-decay scheduler with $5$ warmup epochs. We use the same `mae + cosine similarity' loss~\citep{canturk2024graph}.

The P/SE targets exactly correspond to those chosen in~\citep{canturk2024graph}, viz., $4$ Laplacian eigenvectors (taken in their absolute value) along with corresponding eigenvalues, $7$ Electrostatic PEs, $20$ Random Walk SEs, $20$ Heat Kernel Diag.\ SEs. There is only one exception: we do not consider Cycle Counting Graph Encodings as we observed they would have led to intractable precomputation runtimes in the case of structuralization.

We remark that, when pretraining a model on structuralized graphs, we ask it to predict the above targets on the original nodes calculated for \emph{both} the original and the structuralized form of the input graph. This means having double the pretraining targets than when pretraining a vanilla GPSE model.

\subsection{Downstream evaluation}

In all downstream evaluation experiments we follow the pipeline depicted in \figureref{apd:fig:pipeline}. In particular, component \texttt{(b)} is the pretrained model, from which we extract the generated node representations upstream the prediction heads used in pretraining (\texttt{(c)}). These representations are linearly transformed by component \texttt{(d)}, and then summed along with the encoding of the explicit, original node features in output from component \texttt{(e)}. These constitute the initial node representations updated and later pooled by the downstream GNN depicted in \texttt{(f)}. Component \texttt{(a)} is, effectively, a graph transformation which (i) either discard features or perform structuralization as required; (ii) inject random features as prescribed by the GPSE model~\citep{canturk2024graph}. Note that, during downstream adaptation, only components \texttt{(d)}, \texttt{(e)}, \texttt{(f)} are trained. 

The GNN in \texttt{(f)} is, in all cases, a $3$-layer Graph Isomorphism Network~\citep{xu2019how} employing a single linear pre-message-passing layer and a $2$-Layer Perceptron post-message-passing and before graph readout. The hidden size of this architecture is fixed to $128$. No dropout or weight-decay regularization is applied.

In accordance with~\cite{canturk2024graph}, dropout is applied before and after component \texttt{(d)}. We do not perform tuning of dropout rates, but rather apply the same exact values chosen by the authors in~\citep{canturk2024graph} for the most prominent models in the respective full-dataset adaptation experiments. In particular, for \zinc\, the dropout has a rate of $0.5$ before \texttt{(d)} and is not applied after that component; as for \pept\ the dropout has a rate of $0.1$ before \texttt{(d)} and, again, is not applied thereafter; for \molpcba\ the dropout has a rate of $0.3$ before \texttt{(d)} and $0.1$ after that. We reserve to tune these values for each experiment in future developments of this study; this would allow us to verify whether pretrained representations can be more beneficial in lower data regimes if differently regularized.

As for the optimization approach, we always used Adam~\citep{kingma2014adam} and set a budget of $200$ epochs. As we noticed a potentially significant impact of the optimization scheme on the results, for each experiment we tuned the choice of the base learning rate ($0.0005$, $0.001$, $0.005$) and the scheduler: Reduce-on-validation-plateaux vs.\ Cosine-decay-with-warmup. For the former we set a patience of $10$ epochs; for the latter a number of warmup epochs equal to $5$.

\section{Results on pretraining target predictions}\label{apd:res-pre}

\begin{figure}
    \centering
    \includegraphics[width=0.3\linewidth]{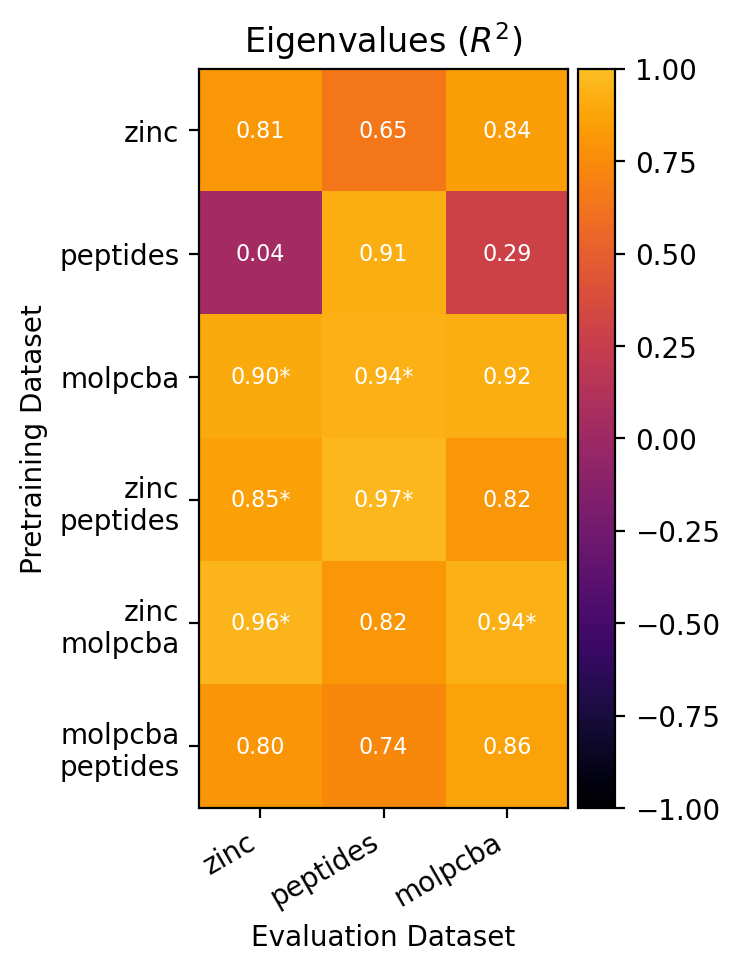}
    \includegraphics[width=0.3\linewidth]{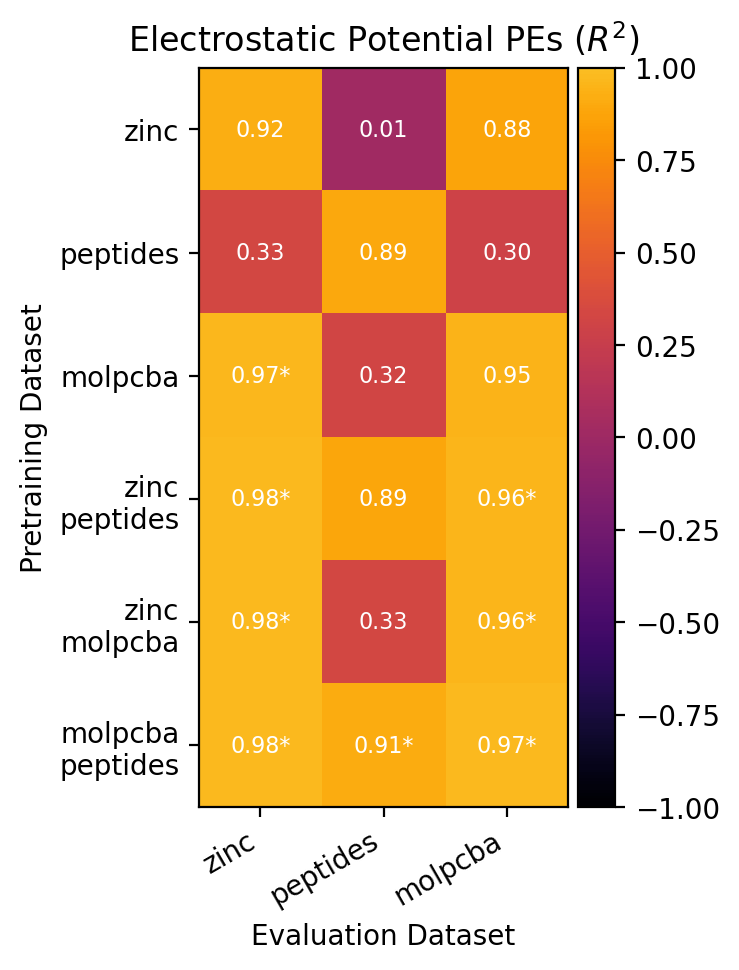}
    \includegraphics[width=0.3\linewidth]{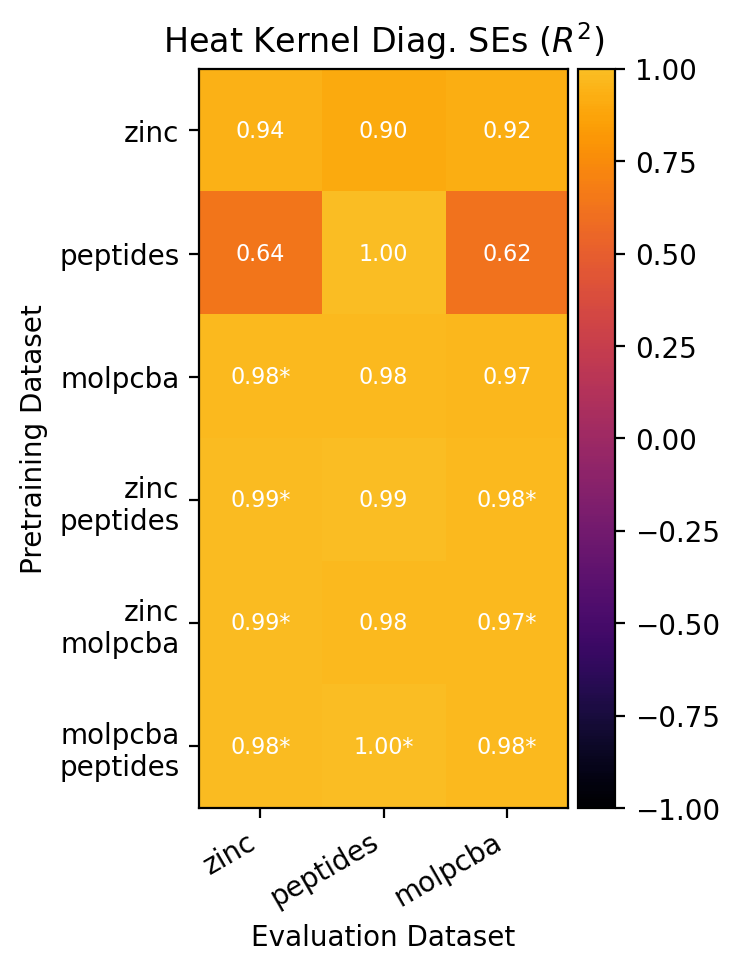}
    \includegraphics[width=0.3\linewidth]{figs//plots/gpse_LapPE.png}
    \includegraphics[width=0.3\linewidth]{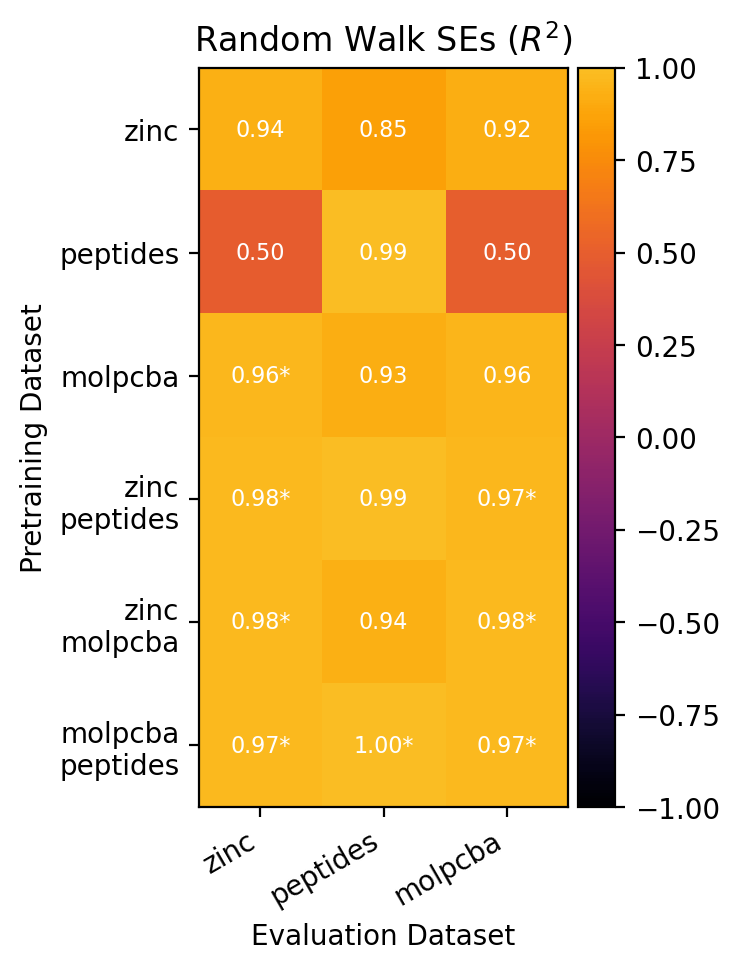}
    \caption{Evaluation of predicting the P/SE targets of the original graphs with GPSE.}
    \label{apd:fig:gpse_pretrain_eval}
\end{figure}

\begin{figure}
    \centering
    \includegraphics[width=0.3\linewidth]{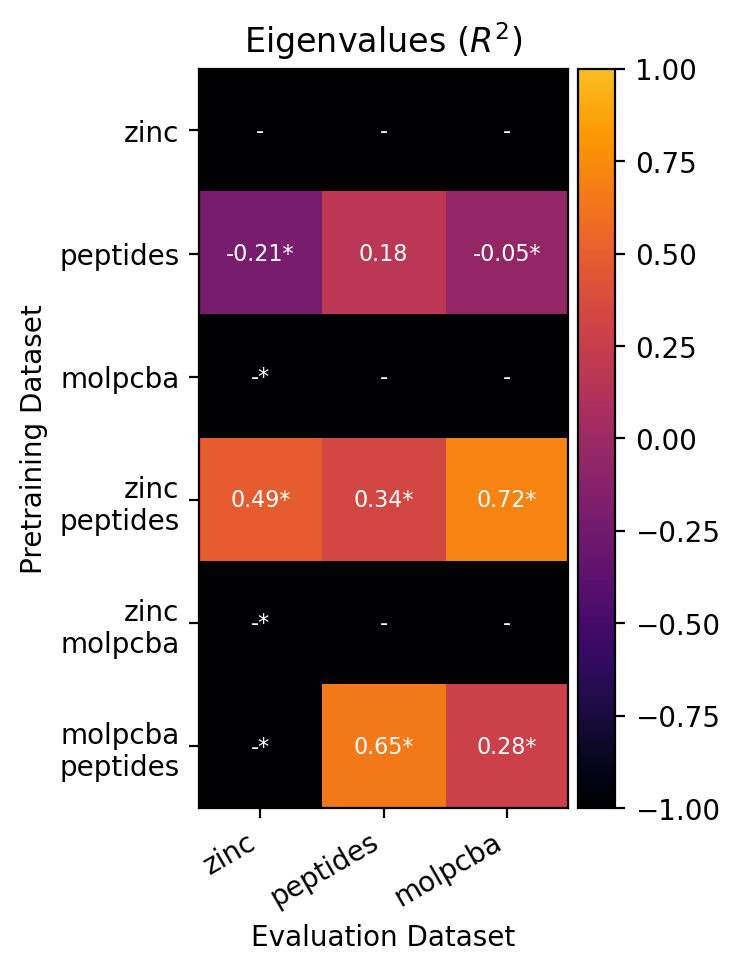}
    \includegraphics[width=0.3\linewidth]{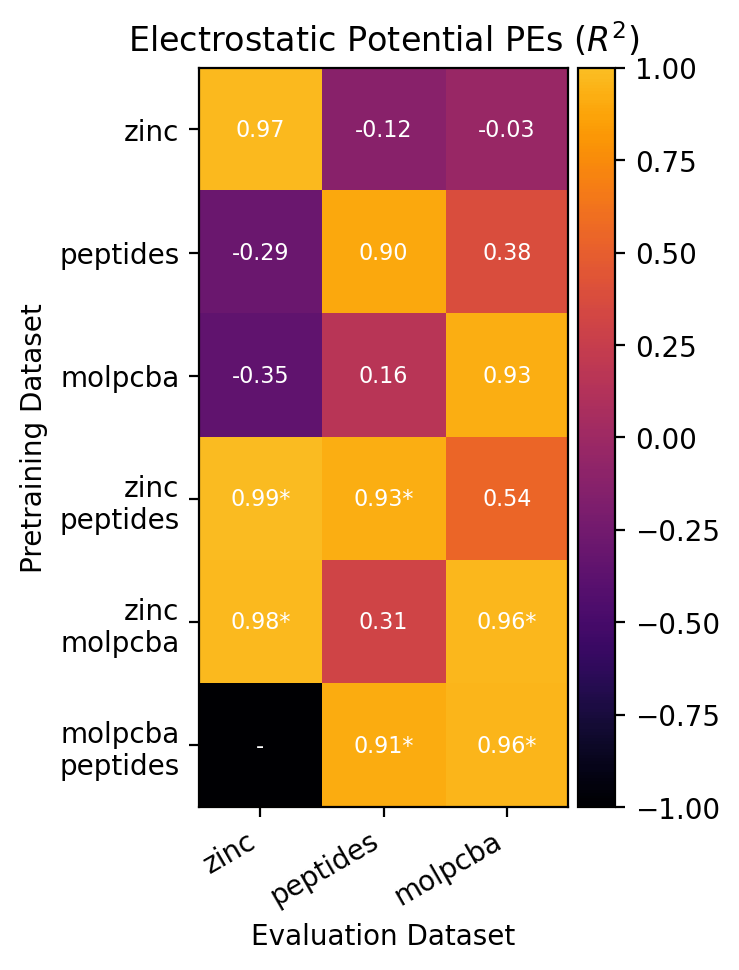}
    \includegraphics[width=0.3\linewidth]{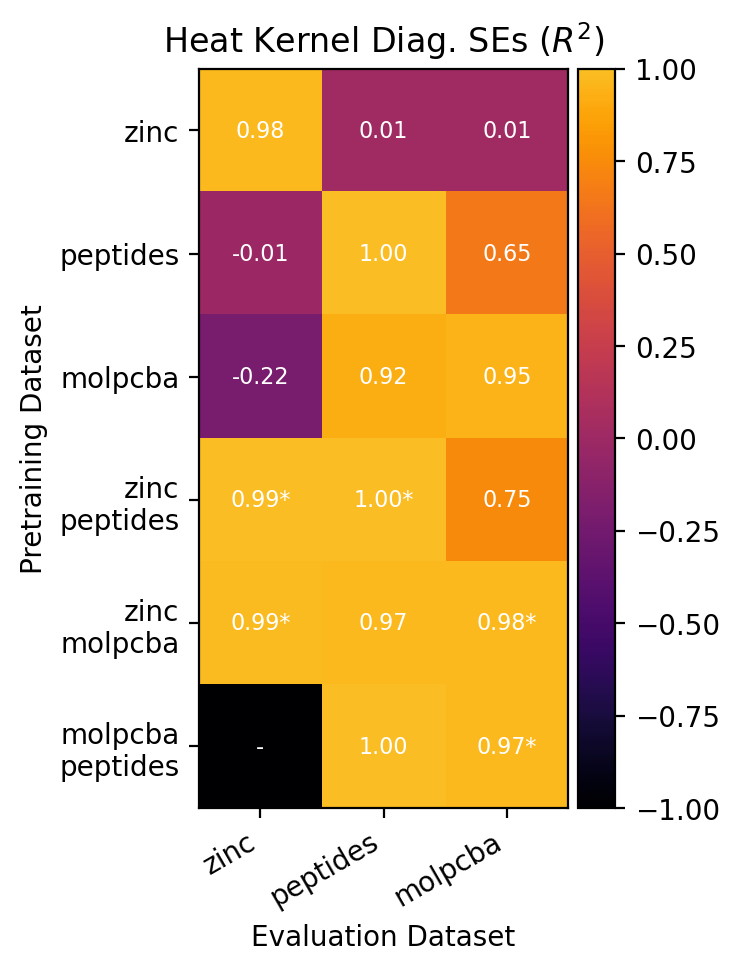}
    \includegraphics[width=0.3\linewidth]{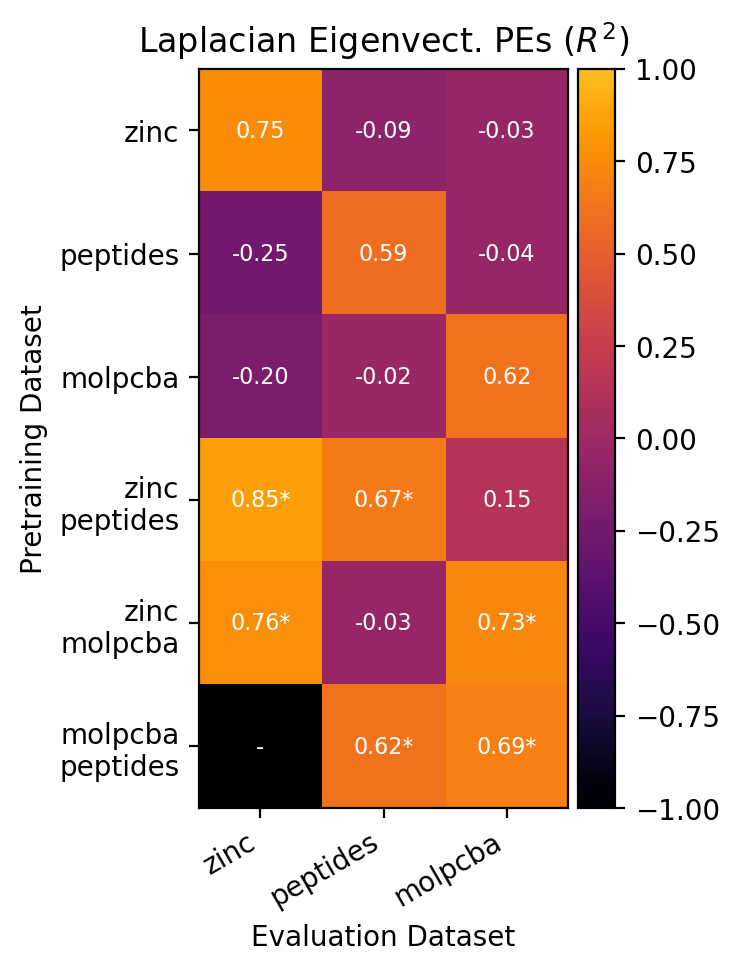}
    \includegraphics[width=0.3\linewidth]{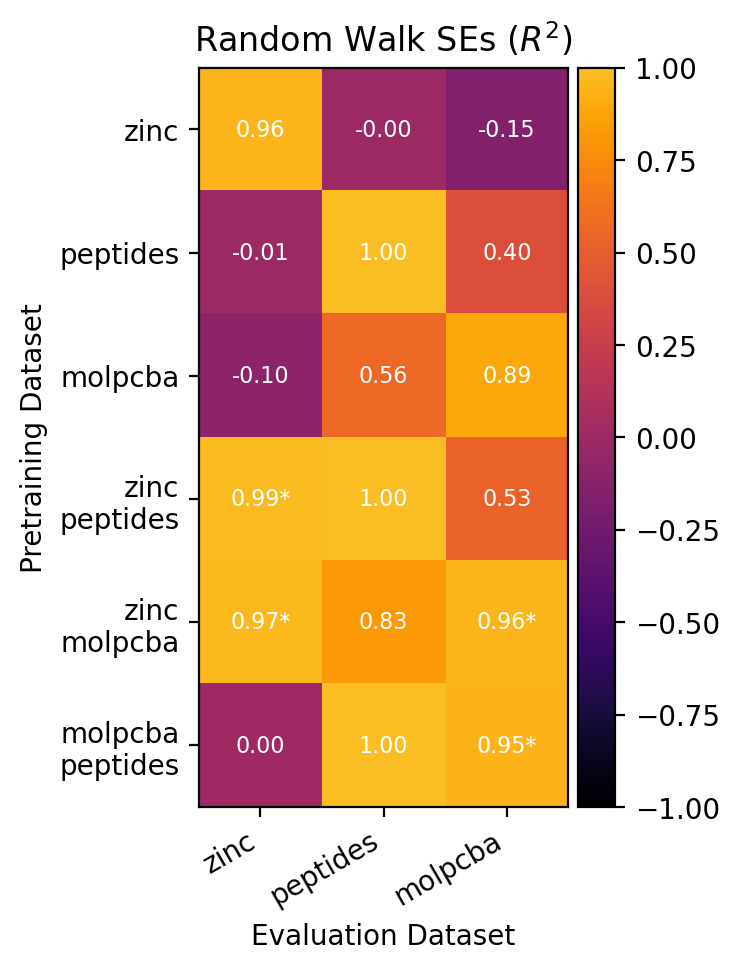}

    \caption{Evaluation of predicting the P/SE targets of the original graphs after structuralization. To ease visualization, negative R$^2$ values are clipped to $-1$, depicted in black (negative values correspond to performance worse than the trivial predictor outputting the mean training target).}
    \label{apd:fig:gpse+struct_pretrain_eval}
\end{figure}

\begin{figure}
    \centering
    \includegraphics[width=0.3\linewidth]{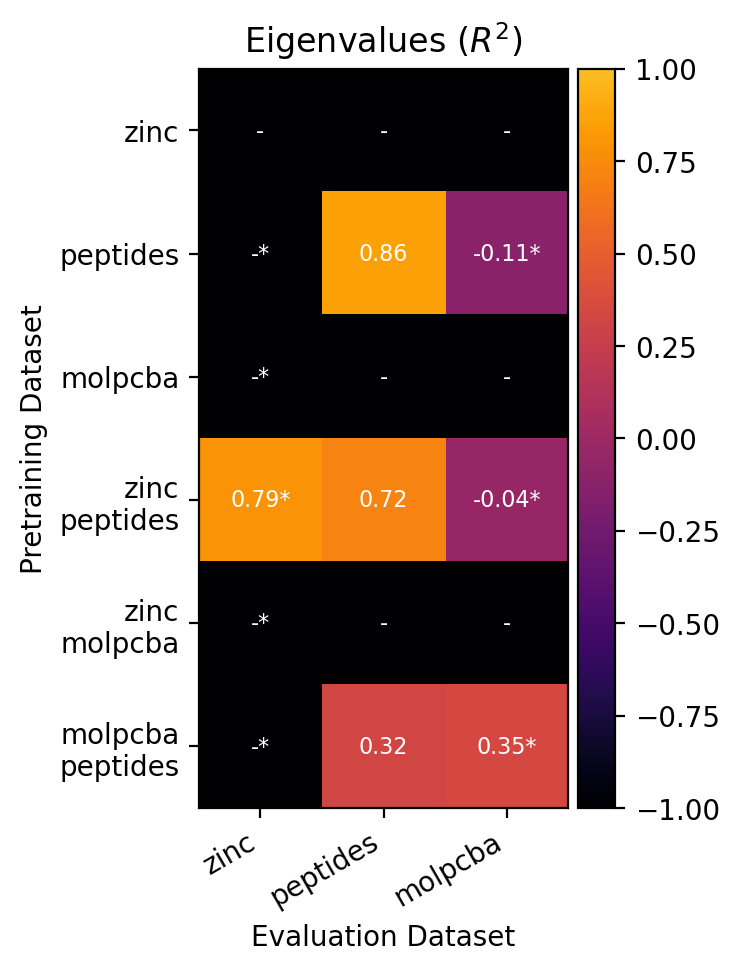}
    \includegraphics[width=0.3\linewidth]{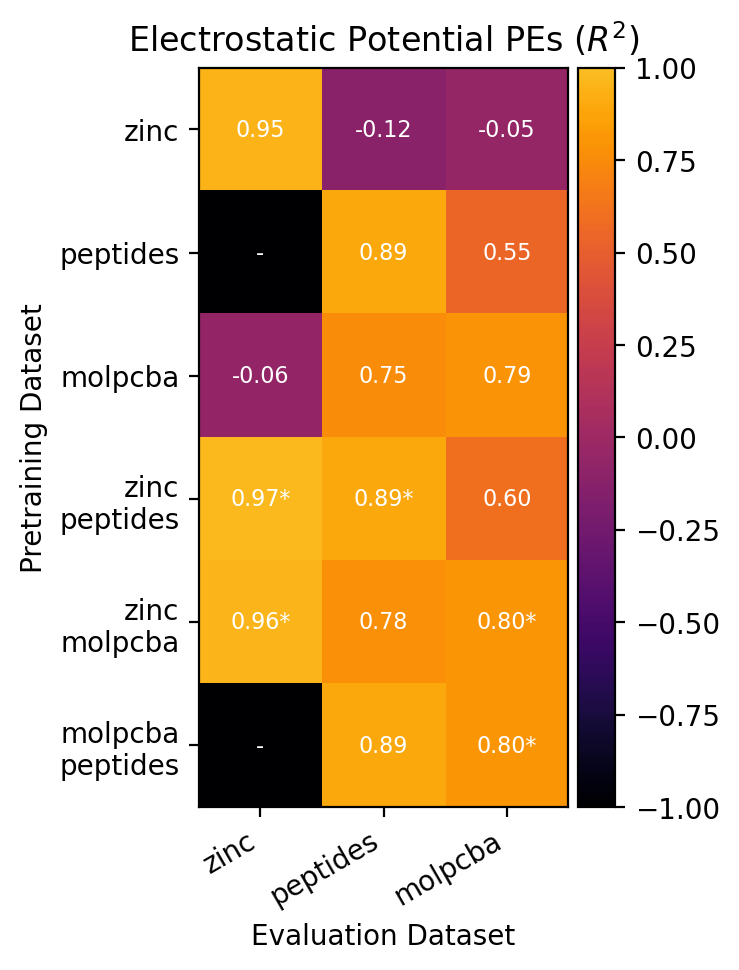}
    \includegraphics[width=0.3\linewidth]{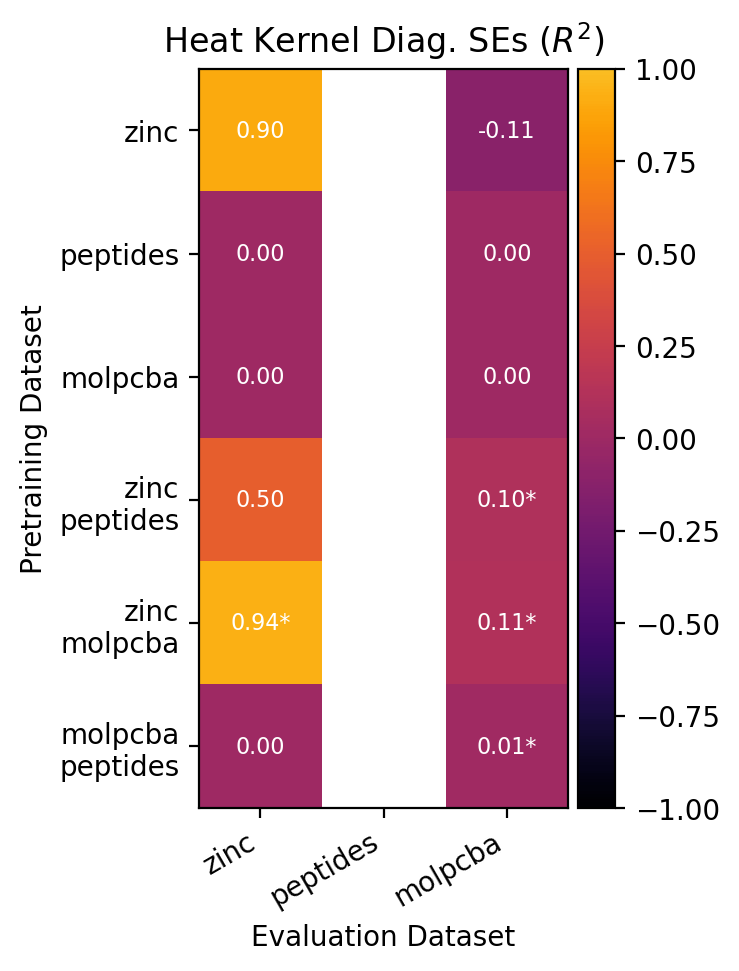}
    \includegraphics[width=0.3\linewidth]{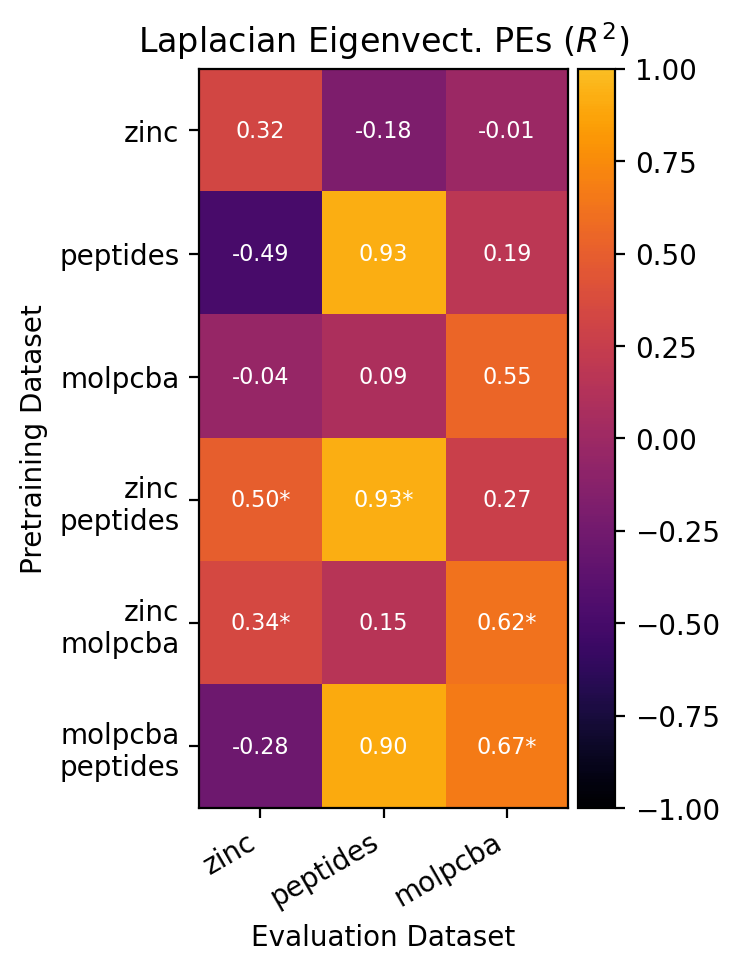}
    \includegraphics[width=0.3\linewidth]{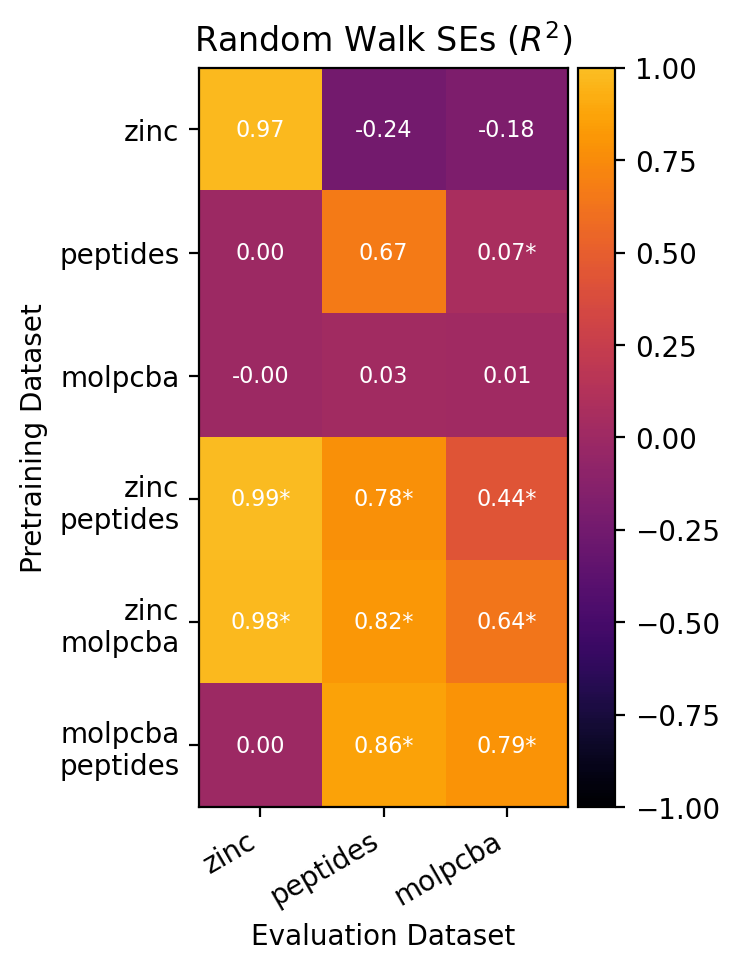}
    \caption{Evaluation of predicting the P/SE targets of structuralized graphs (after structuralization). Test R$^2$ coefficient is reported. To ease visualization, negative R$^2$ values are clipped to $-1$, depicted in black (negative values correspond to performance worse than the trivial predictor outputting the mean training target). White entries are for cases where the calculation of the R$^2$ coefficient yielded non-meaningful numerical values.}
    \label{apd:fig:gpse+struct_pretrain_eval2}
\end{figure}

Figures~\ref{apd:fig:gpse_pretrain_eval}, \ref{apd:fig:gpse+struct_pretrain_eval} and \ref{apd:fig:gpse+struct_pretrain_eval2} show the evaluation of all pretraining targets. In particular, we group P/SEs by category and report the median $R^2$ scores attained on the test sets of the respective evaluation datasets.

\paragraph{Predicting the P/SE targets of the original graphs with GPSE.} We observe that GPSE is able to fit most targets relatively well, although the prediction of Laplacian Eigenvector PEs seems to be a harder task (Figure~\ref{apd:fig:gpse_pretrain_eval}). Either way, we note that achieving satisfactory generalization on \pept{} seems to only be possible if \pept{} was used in the pretraining.

\paragraph{Predicting the P/SE targets of the original graphs after structuralization.} For GPSE on structuralized graphs (Figure~\ref{apd:fig:gpse+struct_pretrain_eval}), we observe our model is only able to generalize to other datasets if pretrained on multiple datasets. This indicates that the inclusion of feature information may render transfer more difficult unless the pretraining corpus is larger and/or `diverse' (and to enquire into the relative impact of size and diversity would be an interesting future endeavor). As for the present results, we find interesting how cross-dataset generalization is relatively better across \molpcba\ and \pept --- which share the same feature encoding --- and is, instead, always problematic when transferring from \molpep\ to \zinc --- which has a different feature encoding. We recall that \emph{targets are purely structural} in this setting and thus hypothesize that the model is harmfully leveraging feature information. One last noteworthy observation is that the prediction of Eigenvalues is particularly hard for this model, even when pretraining `in-dataset'.

\paragraph{Predicting the P/SE targets of structuralized graphs (after structuralization).} Similar conclusions to the above can be drawn for this setting (Figure~\ref{apd:fig:gpse+struct_pretrain_eval2}), except for the fact that the prediction of targets seems generally more difficult --- and sometimes not possible --- even when pretraining `in-dataset' (see, e.g., the prediction of Random Walk SEs on \molpcba{} or Heat Kernel Diag. SEs on all datasets). These generally lower values may contribute to explain the underperformance of structuralization in some downstream evaluation settings, and suggests more focused efforts are required on designing better pretraining strategies and/or constructing better pretraining corpora for our approach. 

\section{Full results on downstream applications}\label{apd:res-down}

\begin{figure}
    \centering
    \includegraphics[width=0.99\linewidth]{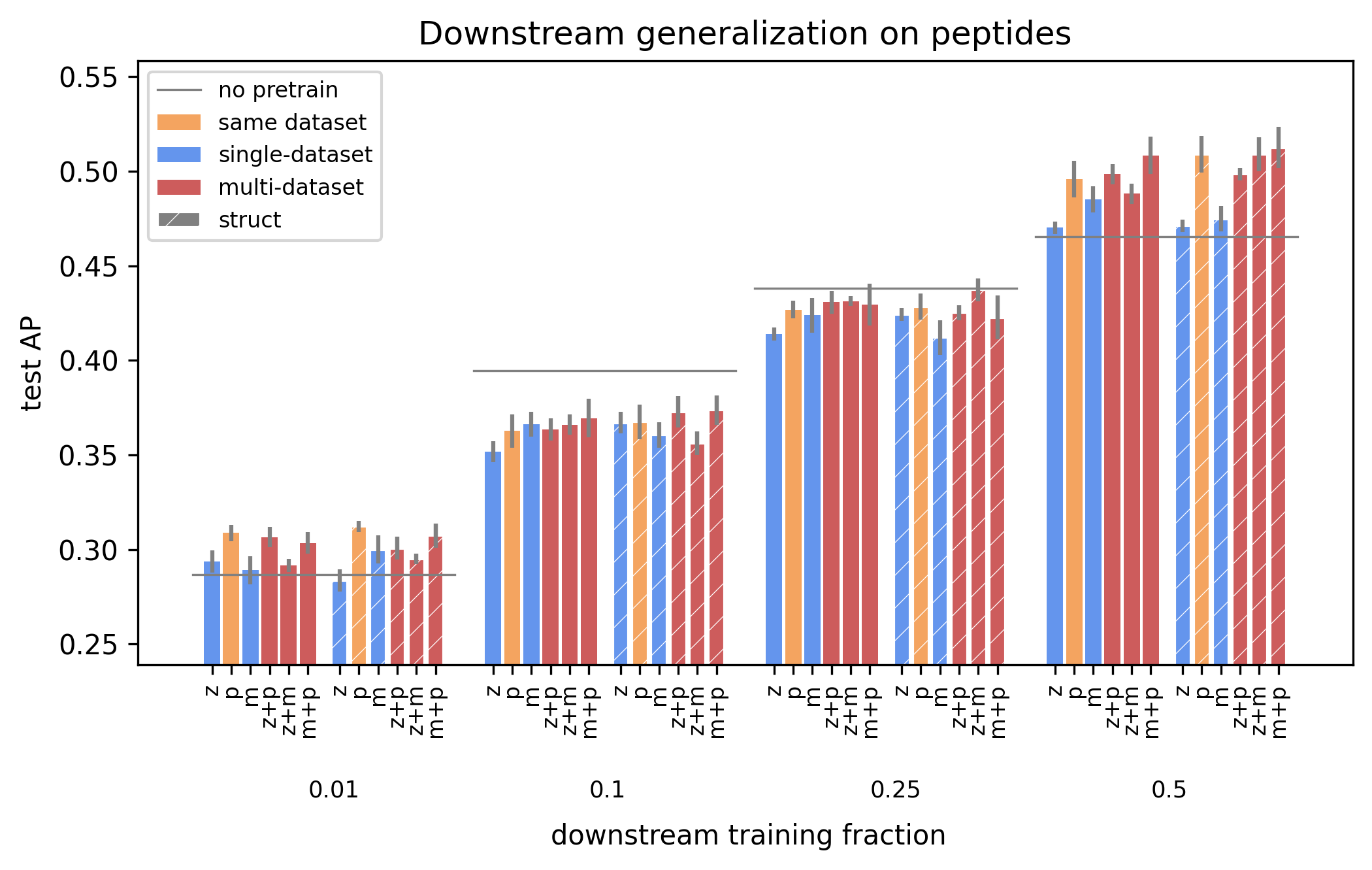}
    \caption{Results of the full downstream evaluation on \pept{}.}
    \label{apd:fig:pep_downstream}
\end{figure}

\begin{figure}
    \centering
    \includegraphics[width=0.99\linewidth]{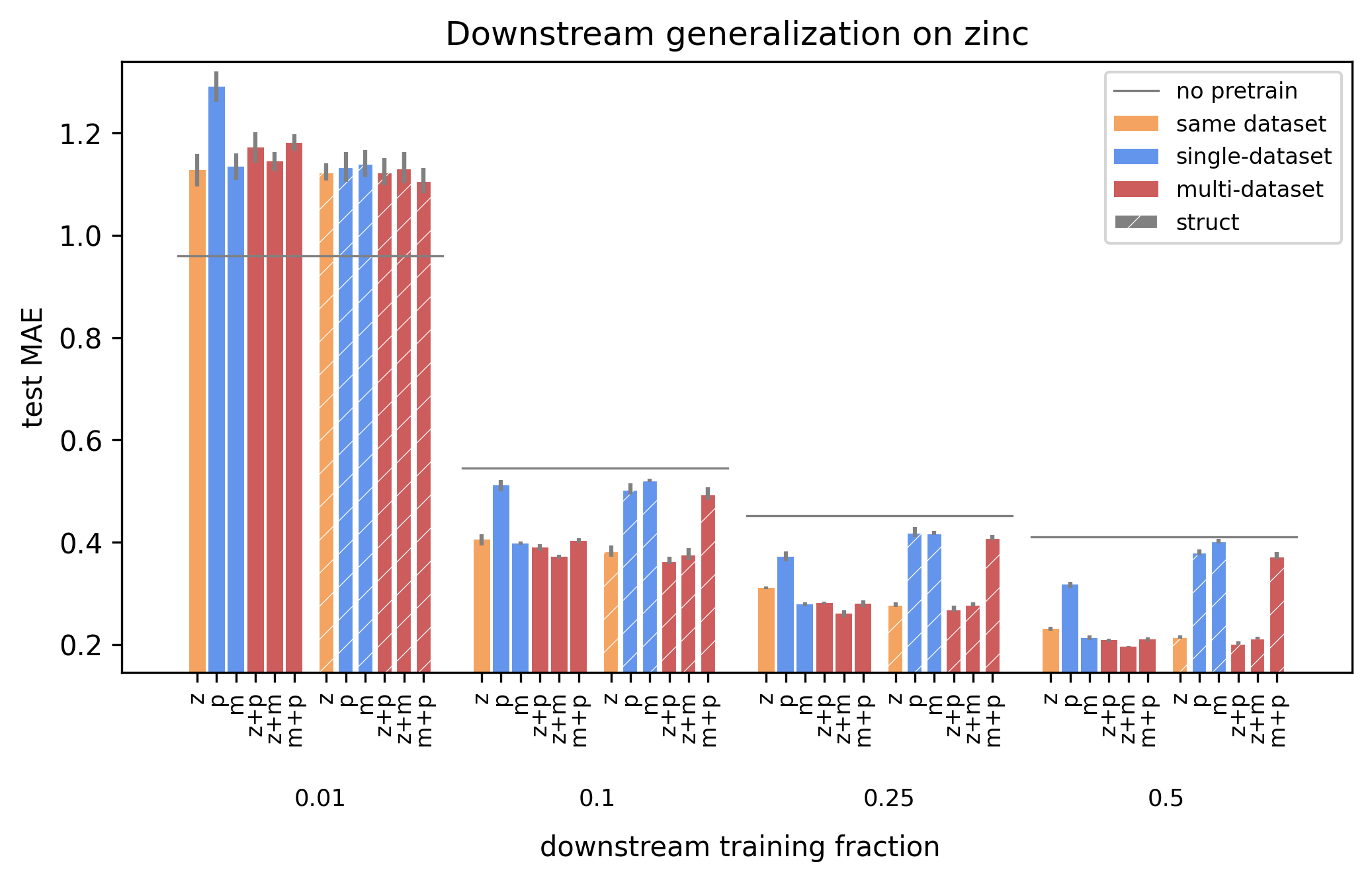}
    \caption{Results of the full downstream evaluation on \zinc{}.}
    \label{apd:fig:zinc_downstream}
\end{figure}

\begin{figure}
    \centering
    \includegraphics[width=0.99\linewidth]{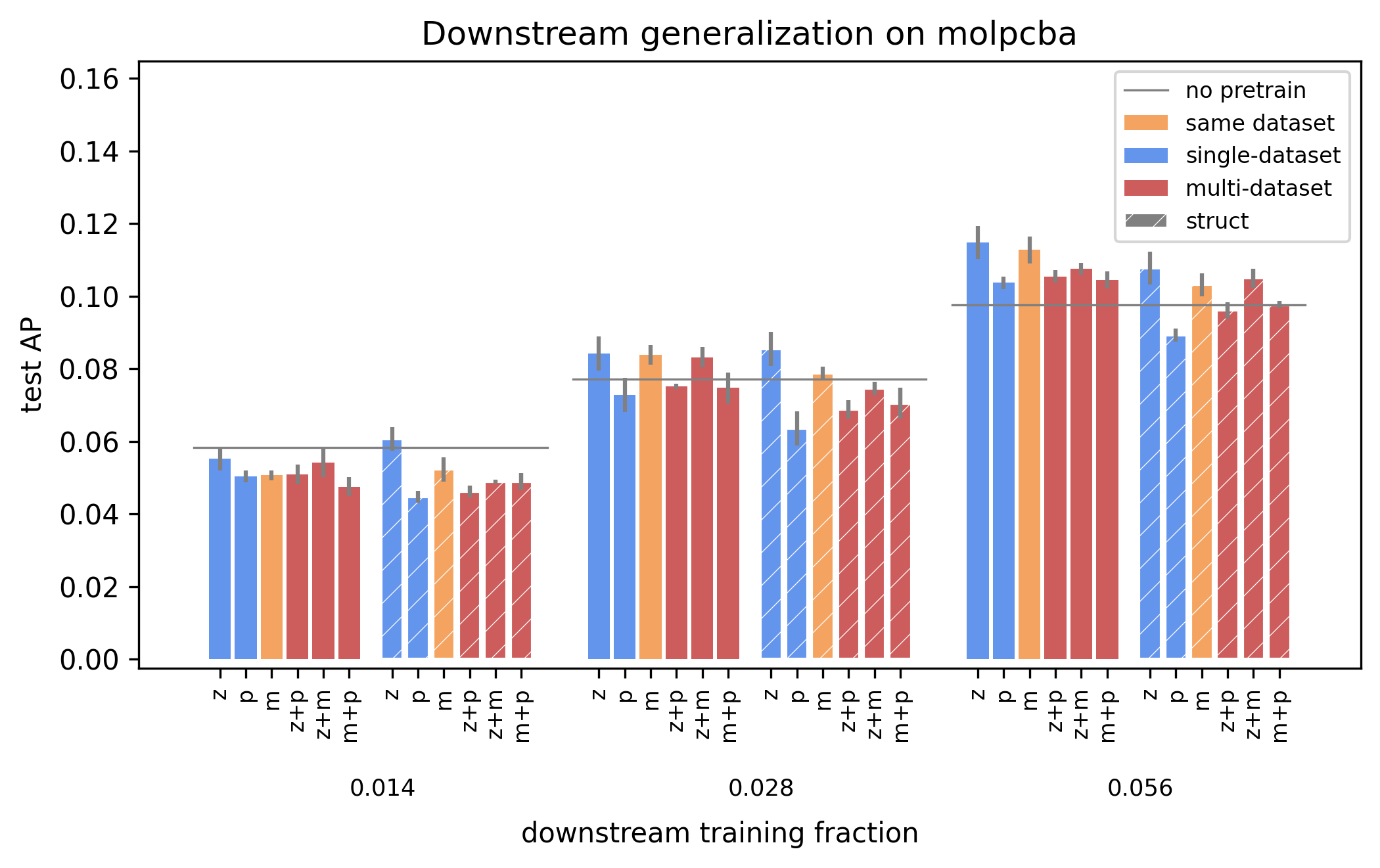}
    \caption{Results of the full downstream evaluation on \molpcba{}.}
    \label{apd:fig:molpcba_downstream}
\end{figure}

We present the full results on all downstream datasets in Figures~\ref{apd:fig:pep_downstream}, \ref{apd:fig:zinc_downstream} and \ref{apd:fig:molpcba_downstream}.

\paragraph{About the lowest data regimes.} In the aforementioned figures it is possible appreciate the behavior of the models in lower data regimes beyond what reported in~\figureref{fig:down}. We observe that node representations from pretrained models are particularly harmful on \pept\ in the $0.1$ training ratio setting, while, intriguingly, they provide more benefits in the lowest data regime ($0.01$ ratio, around $100$ graphs). We hypothesize that, with such data scarcity, on \pept\ it may even be hard to learn a reasonable message-passing scheme, while pretrained embeddings already provide some ready-to-use representations that can more immediately correlate with the targets being predicted. As the amount of training data is increased up to the $0.25$ ratio, the message-passing parameters can more reasonably be fitted (see baseline), but it is likely that the models leverage spurious correlations in the pretrained node embeddings. Finally, for the $0.5$ ratio, enough data are available to optimally make use of both the two sources of information. As for \zinc, the only setting where embeddings from pretrained models are not beneficial is the one of most scarcity, and otherwise provide a significant positive impact. This clearer picture is coherent with the observation that generalizing on \zinc\ becomes relatively easier when the model has access to cyclic-like structural information (as the one that is provided by the pretrained model) and beyond what can be captured by message-passing~\citep{bouritsas2022improving}. 

\paragraph{About \molpcba.} Without even training our models, it is clear that \molpcba{} is a completely different challenge than \zinc{} and \pept{}. This is because (i) the dataset is both significantly larger and (ii) the learning task is seemingly more difficult. For (i), \molpcba{} contains roughly $437,929$ graphs and is thus at least $28$ times larger than \zinc{} ($12,000$ graphs) and \pept{} ($15,535$ graphs). Despite the abundance of data, it seems that it is much more difficult to train a model that generalizes well on \molpcba{}. This is witnessed by the fact that ``strong'' models on \molpcba{} achieve an Average Precision (AP) of $\approx 0.3$ compared to $\approx 0.7$ on \pept{} \citep{canturk2024graph} (ii). Our preliminary results from downstream evaluation on \molpcba\ are reported in \figureref{apd:fig:molpcba_downstream} and indeed showcase the inherent difficulty of the task. Training ratio $0.014$ corresponds to using $\approx 5,000$ training graphs, and is then comparable with the $0.5$ ratio in \zinc\ and \pept. For this amount of data, however, the generalization performance is relatively much worse, as it only hardly reaches $0.06$ test AP. The only regime where GPSE pretrained models seem to be significantly beneficial is ratio $0.056$, four times the size of the training data for which the same pretrained models significantly outperformed the baseline on both \zinc\ and \pept{}. Structuralization does seem to struggle even at this regime, but this may be due to a lower quality of the pretrained models (see~\appendixref{{apd:res-pre}}). In general, we believe that more extensive experimentation is due on \molpcba\ in order to draw more solid conclusions. For example, it may be required to explore substantially larger training ratios, an analysis which we could not carry out at this time with the available resources. 

\section{Using more pretraining data}\label{apd:more-data}

What is the impact of using more pretraining data? Let us first reason on the fact that this is not always possible. In some downstream applications one may only have access to a limited amount of samples, of which only a smaller subset is labeled. In our experiments we have observed that, at least on \zinc{} and \pept{}, it may be beneficial to augment the available in-domain data with additional data sources, and that this sometimes may even improve generalization (see~\figureref{fig:down}).

On the other hand, it is important to notice that, throughout the experiments discussed so far, we have artificially sampled a small fraction of \molpcba's training data to match the size of the training sets in the other datasets. This setting is different that the one considered in~\citep{canturk2024graph}, where models were pretrained on the full set of non-isomorphic structures in \molpcba. A complete analysis on how our full set of results would change when considering the complete \molpcba{} dataset is left for future work; nevertheless, we ran some preliminary experiments we will briefly discuss below.

\paragraph{Pretraining target prediction.} We pretrained two additional GPSE models: one with twice the number of \molpcba{} samples than in the rest of our experiments (around $20$k graphs) (1); one on the full \molpcba{} training set (2). Then, we moved to evaluate the performance of both (1) and (2) on the prediction of pretraining targets across the three datasets. We mention noteworthy results for Electrostatic Potential PEs and Laplacian Eigenvector PEs (the performance on the other targets is mostly saturated already with $10$k pretraining data points). As for the Electrostatic Potential PEs, we report that the performance of GPSE on \pept{} improves from the median $R^2$ of $0.32$ to $0.41$ of (1) and $0.70$ of (2). Concerning the Laplacian Eigenvector PEs, on \pept{}, the median $R^2$ of $0.09$ improves to $0.15$ in the case of (1) and $0.35$ in the case of (2). This suggests that, as already mentioned in~\sectionref{sec:disc}, more data can help in off-domain generalization, although --- as reasonably expected --- much less efficiently than in-domain additional data: $\approx 10$k more (unlabeled) samples from the \pept{}' training set enable median $R^2$'s of $0.91$ and $0.64$ on, resp.\ Electrostatic Potential PEs and Laplacian Eigenvector PEs over the \pept{}' test set. Last we report that the ``in-dataset'' performance for these targets also improve with more data. As for the Laplacian Eigenvector PEs, we report that the performance of GPSE on the \molpcba{}'s test set improves from the median $R^2$ of $0.77$ to $0.85$ of (1) and $0.96$ of (2).

\paragraph{Downstream evaluation on \pept.} We evaluated models (1) and (2) on the \pept{} downstream task, following the same setting and procedure described in~\appendixref{apd:exp-set}. As a first important observation, we note that more pretraining data do not allow models (1), (2) to outperform the baseline in the data regimes $0.1$, $0.25$. We do notice, however, that more pretraining data closes the gap with models pretrained on data mixtures: (1), (2) perform comparably with, e.g., the model pretrained on \zm. Last, we observe that, in the $0.5$ ratio setting, model (2) attains a test AP of $0.513\pm0.002$, slightly outperforming the best pretraining mix \molpep{} ($0.508\pm0.010$).

\end{document}